\documentclass[conference]{IEEEtran}
\IEEEoverridecommandlockouts

\usepackage{cite}
\usepackage{amsmath,amssymb,amsfonts}
\usepackage{algorithmic}
\usepackage{graphicx}
\usepackage{booktabs}
\usepackage{textcomp}
\usepackage{xcolor}

\def\BibTeX{{\rm B\kern-.05em{\sc i\kern-.025em b}\kern-.08em
    T\kern-.1667em\lower.7ex\hbox{E}\kern-.125emX}}
\begin{document}

\title{TAT-VPR: Ternary Adaptive Transformer for Dynamic and Efficient Visual Place Recognition
\\
{}
\author{Oliver Grainge$^{1}$, Michael Milford$^{2}$, Indu Bodala$^{1}$, Sarvapali D. Ramchurn$^{1}$ and Shoaib Ehsan$^{1, 3}$%

\thanks{
This work was supported by the UK Engineering and Physical Sciences Research Council through grants EP/Y009800/1 and EP/V00784X/1}
\thanks{$^{1}$O. Grainge, I. Bodala. S. D. Ramchurn and S. Ehsan are with the School of Electronics and Computer Science, University of Southampton, United Kingdon {\tt\small (email: oeg1n18@soton.ac.uk; i.p.bodala@soton.ac.uk; sdr1@soton.ac.uk; s.ehsan@soton.ac.uk).}}%
\thanks{$^{2}$M. Milford is with the School of Electrical Engineering and Computer Science, Queensland University of Technology, Brisbane, QLD 4000, Australia {\tt\small (email: michael.milford@qut.edu.au).}}%
\thanks{$^{3}$S. Ehsan is also with the School of Computer Science and Electronic Engineering, University of Essex, United Kingdom, {\tt\small (email: sehsan@essex.ac.uk).}}%
}
}

\author{\IEEEauthorblockN{1\textsuperscript{st} Given Name Surname}

\and
\IEEEauthorblockN{2\textsuperscript{nd} Given Name Surname}

\and
\IEEEauthorblockN{3\textsuperscript{rd} Given Name Surname}

}

\maketitle


\begin{abstract}
TAT‑VPR is a ternary‑quantized transformer that brings dynamic accuracy–efficiency trade‑offs to \emph{visual SLAM} loop‑closure. By fusing ternary weights with a learned activation‑sparsity gate, the model can control computation by up to 40\% at run‑time without degrading performance (Recall@1). The proposed two‑stage distillation pipeline preserves descriptor quality, letting it run on micro‑UAV and embedded SLAM stacks while matching state‑of‑the‑art localization accuracy.
\end{abstract}


\section{Introduction \& Background}
Visual Place Recognition (VPR) is often formulated as an image-retrieval task, matching a query image to a geotagged image database. State-of-the-art methods use foundation-scale Vision Transformer (ViT) global descriptors \cite{dinosalad, anyloc, boq}, which are robust to viewpoint, lighting, and seasonal changes. However, their high computational and memory demands limit their use on low-power mobile robots, especially for real-time SLAM loop closure \cite{quantized, pruning}. Consequently, many lightweight SLAM systems still rely on hand-crafted or aggregated point features, sacrificing the robustness of modern transformers \cite{orbslam, kimera}.

To mitigate this, the field has turned to neural network compression. Binary and low-bit quantization can push precision below 8-bit with minimal accuracy loss \cite{binary, floppynet, quantized}, while pruning and sparsity reduce redundancy without degrading performance \cite{pruning}. In NLP, methods like Q-Sparse \cite{qsparse} and BitNet-4.8a \cite{bitnet48} show that sub-4-bit quantization and activation sparsity are compatible. Task-aware knowledge distillation further enhances efficiency by transferring capacity from large ViT teachers to smaller students \cite{lsdnet}. Yet, these approaches are typically fixed to one accuracy-efficiency tradeoff: once deployed, they can’t adapt to changing conditions such as low power or high speeds in cluttered environments.

\begin{figure}[t]
    \includegraphics[width=\columnwidth]{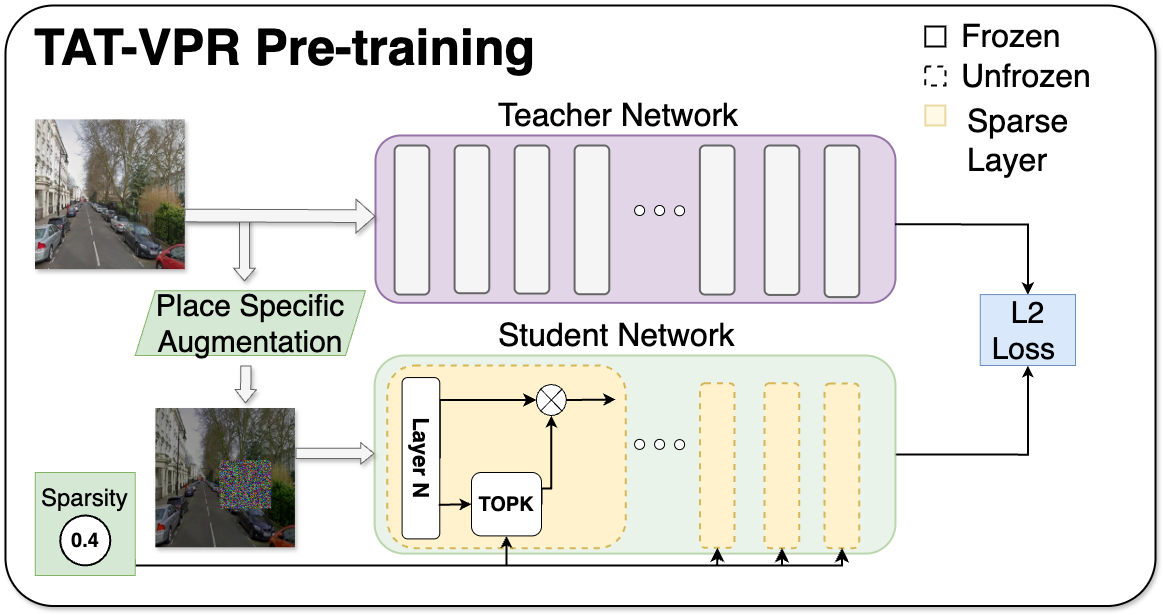}
    \caption{Overview of the TAT-VPR pre-training pipeline. A full-precision DINOv2-BoQ teacher \cite{boq} (purple, frozen) provides token-level supervision to a ternary student transformer (green). During training, the student applies a top-k sparse activation filter. A distillation loss is computed between teacher and student tokens to guide compression-aware representation learning.}
    \label{fig:my_label}
\end{figure}

We propose TAT-VPR, an end-to-end extreme-quantization pipeline (Figure \ref{fig:my_label}) that combines ternary weight quantization, adaptive activation gating, and teacher–student distillation. The result is a model that can dynamically control inference cost by sparsifying activations on demand. On standard VPR benchmarks, TAT-VPR achieves under 1\% Recall@1 drop compared to the dense model while dynamically cutting inference operations by 40\% and shrinking model size 5×, offering a practical, adaptive solution for resource-aware visual localization.

\section{Method}
Our model leverages a ViT-Base architecture for global image–descriptor extraction, modified with ternary weights and controllable sparse activations.  The core components are
outlined below.

\begin{figure*}[!t]
  \centering
  \includegraphics[width=\textwidth]{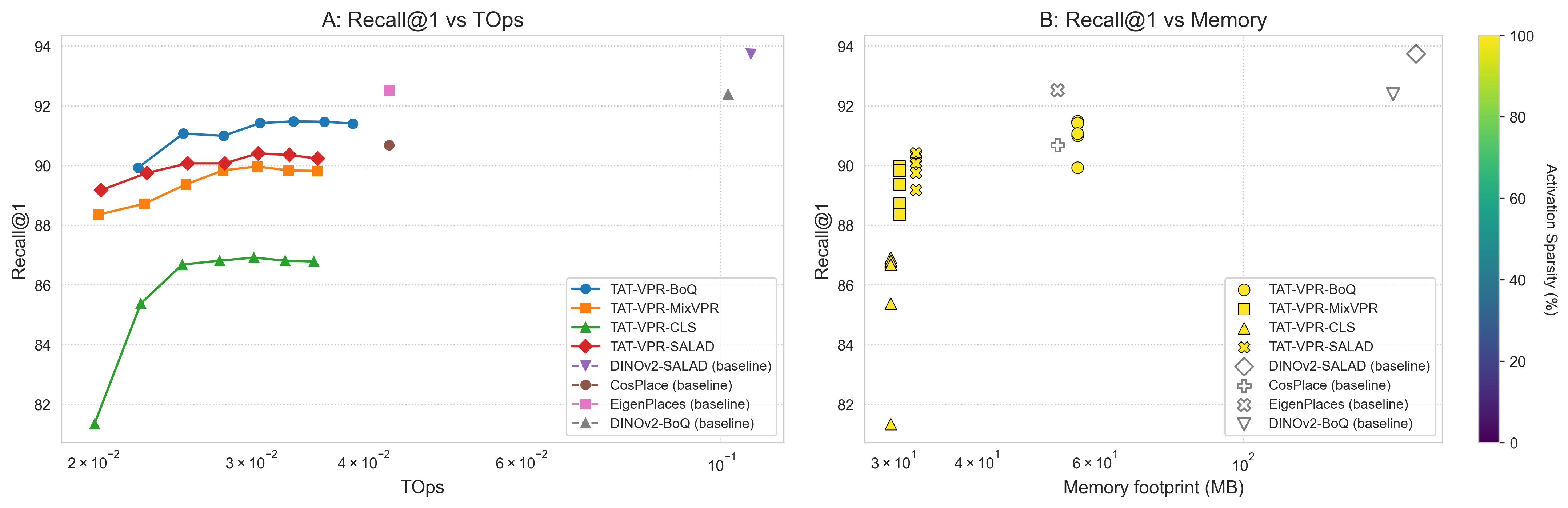}
  \caption{%
    (\textbf{A}) Recall@1 versus Tera-Operations (TOPs) for a feature‐extraction forward pass, showing TAT-VPR curves at activation sparsity levels from 0\% up to 60\%.  
    (\textbf{B}) Recall@1 versus memory footprint on the Pitts30k dataset, highlighting memory savings from ternary-weight backbones.
  }
  \label{fig:combined}
\end{figure*}
\subsection{Ternary Quantized ViT Backbone}
Every weight tensor is quantized to the ternary set \(\{-1,0,+1\}\) with absolute mean quantization:

\begin{equation}
\widetilde{\mathbf W}= \operatorname{RoundClip}\!\Bigl(
\tfrac{\mathbf W}{\gamma+\varepsilon},\,-1,\,1\Bigr)
\label{eq:quant}
\end{equation}

where \(\gamma=\tfrac1{MD}\lVert\mathbf W\rVert_{1}\) is the absolute mean of the tensor and
\(\varepsilon=10^{-6}\) prevents division by zero.
\(\operatorname{RoundClip}\) rounds to the nearest integer then clips to the interval
\([-1,1]\). Resulting in a 8× memory saving compared to 32-bit floating point.

\subsection{Activation-Sparsity Scheduling}
To let the model trade accuracy for efficiency at run time, we apply a
\emph{top-\(k\)} activation mask.  Given activations
\(\mathbf X \in \mathbb R^{N\times D}\),
\begin{equation}
\mathbf M=\operatorname{TopK}\!\bigl(|\mathbf X|,k\bigr),\qquad
\mathbf Y=(\mathbf X\odot\mathbf M)\,\widetilde{\mathbf W}^{\!\top},
\label{eq:topk}
\end{equation}
where the binary mask \(\mathbf M \in \{0,1\}^{N\times D}\) keeps the
largest-magnitude \(k\%\) entries in \(\mathbf X\) and \(\odot\) denotes the hadamard product. Because zeroed elements of \(\mathbf X\) can be
skipped by sparse matrix kernels, only \(k\%\) of the usual
multiply–accumulate operations are executed, yielding proportional
savings in latency, energy, and TOps (Tera Operations).

\subsection{Knowledge Distillation with a BoQ Teacher}
Extreme quantization and sparsity inevitably limit representational capacity, so we pre-train the student with guided distillation from a full-precision DINOv2-BoQ teacher \cite{boq}.  We use a single
token-level mean-squared error loss on the output tokens:
\begin{equation}
\mathcal L_{\text{distill}}=
\frac1{ND}\,
\bigl\lVert\mathbf S^{(L)}-\mathbf T^{(L)}\bigr\rVert_2^{2},
\label{eq:distill}
\end{equation}
where \(\mathbf T^{(L)},\mathbf S^{(L)}\!\in\!\mathbb R^{N\times D}\) are the teacher and student tokens
at the final layer \(L\).  This objective suffices to recuperate the accuracy lost to
ternarisation and sparsity \eqref{eq:topk}. During this stage we linearly raise the sparsity sampling range of \(k\) from \(10\%\) to
\(60\%\), forcing the network to concentrate information in sparse activations whilst avoiding under-training.


\subsection{Fine-tuning}
We fine-tune the pre-trained compact backbone on the \textsc{GSV-Cities} dataset using a supervised multi-similarity retrieval loss \cite{gsvcities}. Four aggregation heads are evaluated on top of the frozen sparse ternary backbone: Bag-of-learnable queries, SALAD, MixVPR, and a lightweight classification token head \cite{boq, dinosalad, mixvpr}. Only the head and the last two backbone layers are updated to preserve low-precision representations and prevent overfitting.



\begin{table}[!t]
  \centering
  \scriptsize 
  \caption{Recall@1 / Recall@1 per MB on SVOX condition splits.}
  \label{tab:tab1}
  \resizebox{\columnwidth}{!}{%
    \begin{tabular}{lccccc}
      \toprule
      \textbf{Method}        & \textbf{Snow}    & \textbf{Rain}    & \textbf{Overcast} & \textbf{Night}   & \textbf{Sun}     \\
      \midrule
      TAT-BoQ       & 97.0 / 1.71 & 94.2 / 1.66 & 97.7 / 1.72 & 61.5 / 1.08 & 92.7 / 1.63 \\
      TAT-MixVPR    & 90.6 / \textbf{2.95} & 87.5 / \textbf{2.85} & 93.5 / \textbf{3.04} & 35.2 / 1.14 & 82.2 / \textbf{2.67} \\
      TAT-CLS       & 78.7 / 2.63 & 75.2 / 2.52 & 89.7 / 3.00 & 20.0 / 0.67 & 66.5 / 2.23 \\
      TAT-SALAD     & 95.3 / 2.93 & 92.5 / 2.84 & 96.7 / 2.97 & 41.6 / \textbf{1.28} & 88.1 / 2.71 \\
    \midrule      
  \addlinespace[0.3em]  
      DINOv2-SALAD  & \textbf{99.4} / 0.30 & \textbf{98.7} / 0.29 & \textbf{98.5} / 0.29 & \textbf{97.8} / 0.29 & \textbf{97.7} / 0.29 \\ 
      DINOv2-BoQ    & 98.7 / 0.27 & 98.5 / 0.27 & 98.3 / 0.27 & 95.4 / 0.26 & 97.1 / 0.27 \\
      CosPlace      & 90.3 / 0.85 & 85.1 / 0.80 & 91.4 / 0.86 & 48.6 / 0.46 & 76.9 / 0.73 \\
      EigenPlaces   & 91.5 / 0.86 & 88.0 / 0.83 & 92.5 / 0.87 & 59.8 / 0.56 & 85.2 / 0.80 \\
      \bottomrule
    \end{tabular}%
  }
\end{table}

\section{Results and Discussion}

Figure~\ref{fig:combined}-A shows the Recall@1 versus computational cost (in TOps) curves for TAT-VPR, where computational cost is controlled via runtime activation sparsity. Across all aggregation heads, up to a 40\% reduction in TOps can be achieved with less than a 1\% loss in Recall@1 on the Pitts30k dataset. In contrast, baseline models including \cite{dinosalad, boq, cosplace, eigenplaces} rely on a fixed and higher volume of TOps.

Figure~\ref{fig:combined}-B presents the static memory consumption of TAT models compared to baselines. Despite employing larger backbones in terms of parameter count, most TAT models (excluding TAT-BoQ) consume significantly less memory. This efficiency arises from the use of 2-bit ternary quantized weights in the TAT models.

Table \ref{tab:tab1} demonstrates the robustness of TAT models relative to baselines under various appearance change conditions. Due to the distillation of generalizable representations during the pre-training stage, both TAT-BoQ and TAT-SALAD achieve higher recall scores than convolutional baselines such as \cite{cosplace, eigenplaces}, even when operating at 40\% activation sparsity. Moreover, across all datasets, TAT models achieve markedly superior memory efficiency despite the 40\% activation sparsity constraint.

\section{Conclusion}
TAT‑VPR brings dynamic scalability to \emph{visual SLAM}: its ternary weights and adaptive activation sparsity let a single network down‑shift latency and power on the fly, yet still deliver near‑state‑of‑the‑art Recall@1 for loop‑closure detection.  The 5× memory cut and 40\% TOps savings free headroom for tracking, mapping, and relocalisation threads on micro‑UAV and embedded SLAM stacks.

\bibliographystyle{IEEEtran}
\bibliography{references}

\vspace{12pt}

\end{document}